\documentclass[conference]{IEEEtran}
\IEEEoverridecommandlockouts

\usepackage{cite}
\usepackage{amsmath,amssymb,amsfonts}
\usepackage{graphicx}
\usepackage{comment}
\usepackage{textcomp}
\usepackage{xcolor}
\usepackage{hyperref}
\usepackage{float}
\usepackage{tabularx}
\usepackage{gensymb}
\usepackage{siunitx}
\usepackage{booktabs}
\usepackage{multirow}
\usepackage{array}
\usepackage[caption=false,font=footnotesize]{subfig}
\usepackage{algorithm}
\usepackage{algorithmic}
\usepackage[T1]{fontenc}
\usepackage[utf8]{inputenc}

\begin{document}

\title{
HyDRA Scorpion: A Cost-effective and Modular ROV for Real-Time Underwater Inspection, Intervention, and Object Detection
}

\author{
\IEEEauthorblockN{
Anika Tabassum Orchi$^{1}$,
Md Farhan Zaman$^{1}$,
Md Darain Khan$^{1}$,
Md Alamgir$^{2}$,
Mahbubul Islam$^{1}$,\\
Md. Jobayer Rahman$^{1}$,
A. M. Zayed Abdullah$^{1}$,
Md Mehrab Hossain Khan$^{1}$,
Md. Kutub Al Baki$^{2}$,
Iftekharul Islam$^{1}$,\\
Shakil Ahmed$^{1}$,
Md Sadique Hossain$^{1}$,
Md Muzahidul Islam$^{1}$,
Shah Mohammad Seaman$^{1}$,
Nusrat Jahan Piyal$^{1}$,\\
Shekh Md. Saifur Rahman$^{1}$,
Fahim Hafiz$^{1}$,
A.K.M. Muzahidul Islam$^{1}$\thanks{Corresponding author. Email: muzahid@cse.uiu.ac.bd},
M. Rezwan Khan$^{2}$
}

\IEEEauthorblockA{
$^{1}$Department of Computer Science and Engineering,\\
United International University, United City, Madani Ave, Dhaka 1212, Bangladesh\\
}

\IEEEauthorblockA{
$^{2}$Department of Electrical and Electronic Engineering,\\
United International University, United City, Madani Ave, Dhaka 1212, Bangladesh\\
}
}

\maketitle

\vspace{-1em}


\begin{abstract}
A Remotely Operated Vehicle (ROV) is a tethered underwater robot used for tasks like inspection and intervention. While essential tools for underwater science, the high cost of commercial ROVs and a persistent gap between mechanically capable platforms and those with integrated intelligence create a significant barrier to access. HyDRA Scorpion differs from conventional systems by addressing these challenges, integrating an advanced, AI-driven perception stack with in-situ measurement capabilities onto a low-cost, locally manufacturable platform. The system combines 4-DoF maneuverability, dual manipulators, and a custom pressure-tested housing. Experimental results validate the system's robustness and performance. Leak-free operation was confirmed through prolonged pressure testing of the electronics housing to 4 bar, equivalent to the pressure of a 304.8-meter water depth approximately in a simulated environment, with no moisture ingress detected. The vehicle also demonstrated stable station-keeping, maintaining its position within a tight tolerance of \(\pm\)0.15 meters under external disturbances. The onboard AI module achieved underwater object detection mean Average Precision (mAP) of 0.89 with real-time inference, length and 3D-mapping based distance measurement. Also, 4-DoF manipulator arm can grip and maintain dual-function manipulator feature which support 360 degree tangle-free rotation. 

\end{abstract}


\section{Introduction}
Marine robot technologies have progressed with new advancements in discovering and monitoring underwater environments with high accuracy and efficiency~\cite{zhou2023review}. One of the key technologies among these is the Remotely Operated Vehicle (ROV)~\cite{yoerger2021hybrid}, which eases scientific investigation~\cite{vedachalam2019autonomous}, offshore operations, and underwater inspection~\cite{jorge2021samplingEstimation}. Riverine nations, with long coastlines and rich marine resources~\cite{hasan2024oceanic}, needs indigenous, low-cost, and reliable underwater investigation and surveillance tools. Scuba diving remains hazardous to human beings~\cite{lambert2024underwater}, and foreign-bought ROVs are costly and not well-suited to local conditions~\cite{gerungan2020scuba}. Therefore, a cutting-edge, cost-effective, environment-friendly, and modular robotic platform is necessary to sustain the research needs. The Hydra Scorpion project was motivated by the need for an indigenous, low-cost ROV tailored to the marine conditions, aiming to support research, search and rescue efforts, and the sustainable monitoring of seabed resources.

\begin{figure}[ht] 
\centering 
\includegraphics[width=0.45\textwidth]{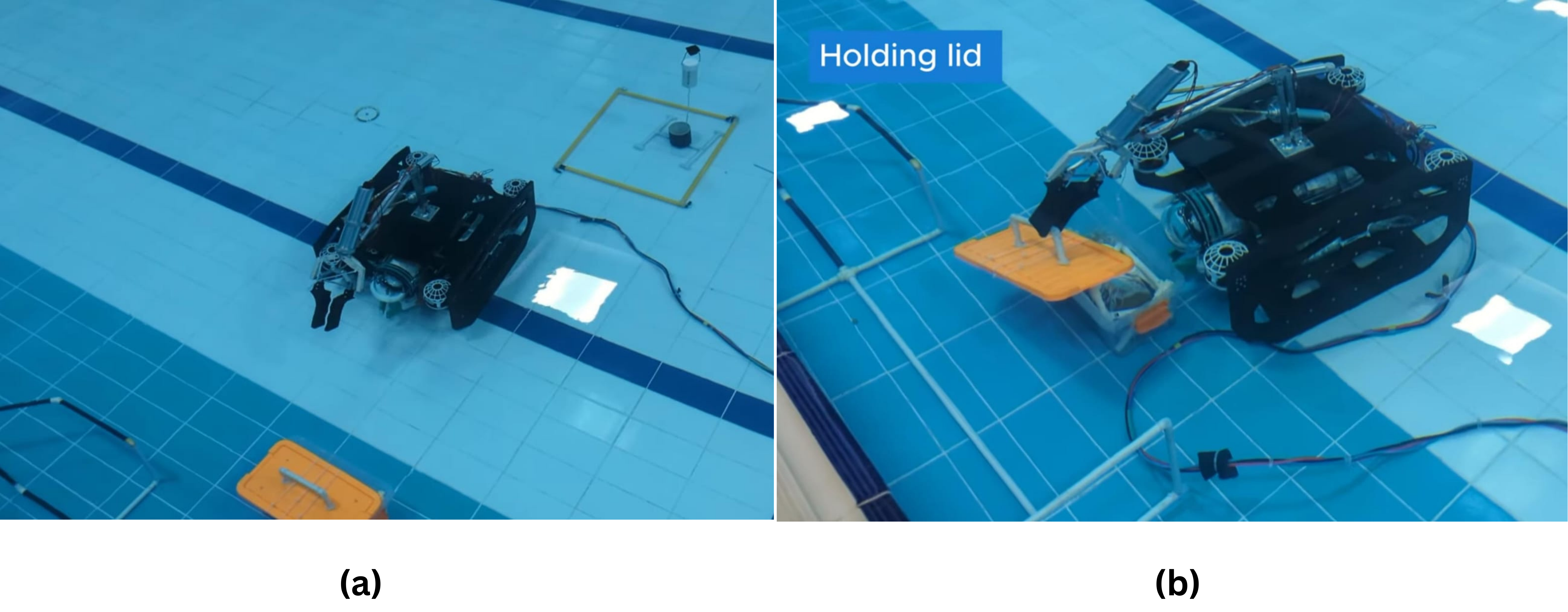} 
\vspace{-1em}
\caption{(a) HyDRA Scorpion ROV navigates by leveraging the AI-based guidance from the in-vehicle module and its down-facing camera, and (b) 3D environmental mapping and intervention}
\label{fig:rov_operating}
\end{figure}

Underwater robotics underpins ocean science, offshore energy, aquaculture, and infrastructure maintenance. However, many institutions in developing countries face a persistent barrier: the high cost and dependence on imports. High-quality commercial ROVs provide reliability and payload capacity, but their price, logistics, and proprietary ecosystems restrict access to hands-on research and field deployment. In the case of Low- and Middle-Income Countries (LMICs), mostly costly foreign ROVs and traditional methods are used, which do not perform well in muddy water, provide poor measurements, and are not accessible to students. But these remedies are inadequate, and they cannot compensate for local conditions; they lack 3D mapping~\cite{ferrera2021hyperspectral}, and they’re too expensive to deploy everywhere. Hydra Scorpion aims to address this with an accessible, AI-driven platform that can accurately identify, measure, and map water-based objects. The current ROVs are physically capable, but do not have the intelligence~\cite{neira2021review}; image processing is generally done as simple demos without any real detection or measurement~\cite{pinto2020maresye}. HyDRA Scorpion bridges this gap with an affordable, modular design and an end-to-end perception stack that merges deep learning, classical vision, and a proprietary embedded metrology stack to provide genuine inspection capability. To address these challenges, we have developed an advanced Remotely Operated Vehicle named HyDRA Scorpion. This ROV is designed to conduct sophisticated underwater operations, including real-time object detection in turbid water, AI-assisted length and distance measurement, 3D environmental mapping, and complex manipulation with a 4-DOF robotic arm. The vehicle's core operational concept, combining autonomous navigation with mapping and intervention, is illustrated in Figure~\ref{fig:rov_operating}.

The system integrates sophisticated hardware, real-time video streaming, and innovative software with potential for use in real-world marine investigations. The overall system is equipped with an AI-driven (YOLO + OpenCV) camera module for object detection and tracking in diversified environments. We utilized AI-based algorithms for the accurate measurement of objects or wreck sizes, recording visual information, replicating high-fidelity 3D underwater scene models, and computing precise vehicle-to-object distances. The ROV features a manipulator arm with several degrees of motion for grasping objects, along with trusted thrusters for stable operation and accurate control. We designed the arm by using PyQt6, with joystick input, live camera display, and 3D visualization. The hardware was developed using a Raspberry Pi–Pixhawk 4 with motor drivers and feedback systems. The principal contributions of this research are the following:

\begin{itemize}
\item We developed a low-cost, high-performance ROV for complex underwater operations.

\item We integrated real-time video streaming with AI-based object detection and length measurement functionality.

\item We developed precise systems for 3D mapping and measuring distances.

\item We designed a 4-DOF robotic arm for underwater manipulative and retrieval tasks.

\end{itemize}

The rest of the article is illustrated as follows: Section \ref{lr} provides a summary of the existing literature. Section \ref{sdi} presents an overview of the whole System design and implementation. In Section \ref{ra}, the experimental results and outputs are described. Finally, we conclude our work and discuss some future research directions in Section \ref{con}.

\section{Related Work}\label{lr}

\subsection{Evaluation of System design}

\begin{figure}[ht] 
\centering 
\includegraphics[width=0.50\textwidth]{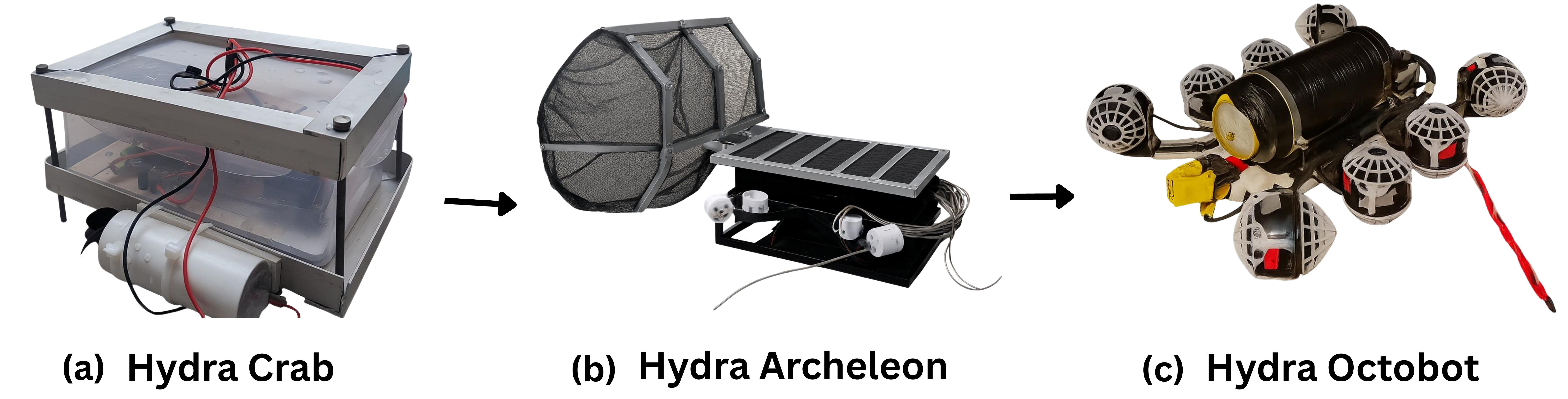}
\vspace{-1.5em}
\caption{Chronological Evaluation of Hydra Family, (a) Hydra crab; (b) Hydra Archeleon; (c) Hydra Octobot}
\label{fig:Hydra_Family}
\end{figure}

\begin{figure}[ht] 
\centering 
\includegraphics[width=0.50\textwidth]{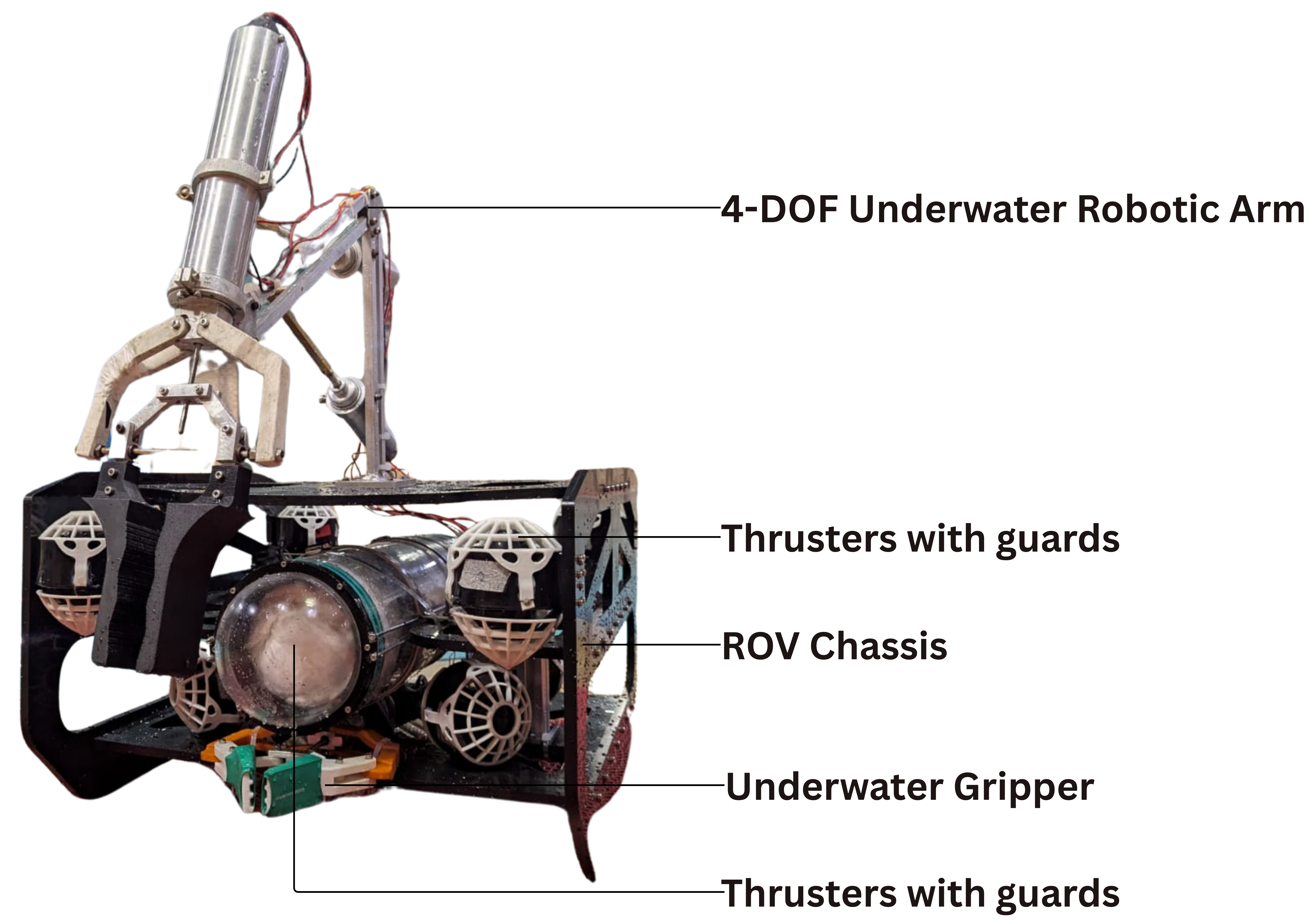} 
\vspace{-1.5em}
\caption{The prototype of HyDRA Scorpion}
\label{fig:prototype}
\end{figure}

The Hydra series presents four innovative aquatic robotic platforms, each tailored for distinct missions. The Hydra Crab is a lightweight (2 kg) wireless RC surface vehicle equipped with two thrusters, ideal for smooth navigation on surface waters. Designed for deeper operations, the Hydra Archeleon is a robust submersible ROV weighing 15 kg, featuring 6–8 thrusters and capable of reaching depths of up to 30 m (100 ft), making it suitable for inspection and exploration tasks. The Hydra Octobot is a compact 6-DoF submersible powered by eight thrusters, featuring a streamlined 6 kg frame and an impressive depth capability of 60 m (200 ft), which offers high maneuverability for precision underwater missions. Finally, we have developed Hydra Scorpion, a weighted 17.6 kg vehicle with eight thrusters, capable of diving depths of up to 1000 ft, tested under pressure in a simulated environment. Figure \ref{fig:Hydra_Family} displays the chronological development of the Hydra family, and Figure \ref{fig:prototype} represents the final prototype of HyDRA Scorpion.

\subsection{Related Work}
Research on Remotely Operated Vehicles (ROVs) has traditionally focused on platform design, actuation, and control strategies. Several studies present the development of six-degree-of-freedom vehicles with novel thrust allocation and experimental validation. These works demonstrate strong mechanical and control architectures, such as cost-effective 6-DoF prototypes and thrust-distribution methods for enhanced maneuverability~\cite{martin2016fully, salem2023design}.

Beyond vehicle platforms, recent work~\cite{lin2025odyssee, grimaldi2502stonefish} on subsea autonomy emphasises semantic mapping and intervention. Semantic mapping frameworks provide higher-level situational awareness and task reasoning, but they presuppose reliable object detection and quantitative measurements as inputs. Similarly, efforts in underwater exploration ROVs and search-and-rescue prototypes highlight mission versatility yet do not provide specialized vision pipelines~\cite{kabanov2021design} for damage identification or metrological assessment. Complementary efforts in simulation and dataset development have also supported the field~\cite{grimaldi2502stonefish}. The Stonefish simulator has enabled machine learning workflows in marine robotics but remains constrained by the sim-to-real gap, particularly under challenging optical conditions such as turbidity, backscatter, and biofouling. Public underwater datasets, such as those developed for the URPC and DUO challenges, have standardized comparisons for biota detection tasks, including holothurians, echinus, scallops, and starfish. While invaluable, these datasets are narrowly scoped toward aquaculture monitoring and lack structural defect classes or ground truth suitable for inspection missions~\cite{fan2020dual}. Rapid progress has been made in adapting object detection models like YOLO and Faster R-CNN for underwater~\cite{wen2024enyolo, zhao2025lightweight}. These models are now better at handling noise and detecting small objects, making them suitable for real-time use. However, most of these systems are only tested on marine animals or generic objects and don't provide specific information like damage analysis or measurements~\cite{zhang2021lightweight, zhao2022improved}. Also, most are just algorithmic demonstrations and not fully integrated into a complete remotely operated vehicle (ROV).

While these platforms are technically robust, they remain limited in perception capabilities~\cite{han2020underwater}. Image processing, when implemented, is typically restricted to basic demonstrations without trained detection models, real-time deployment, or quantitative measurements of structural features~\cite{islam2020fast, pinto2020maresye}. Consequently, the gap between mechanically capable ROV platforms and inspection-grade perception remains significant. HyDRA Scorpion overcomes these limitations by using a YOLOv6 model with classic image processing to detect cracks and structural damage in challenging underwater conditions. It goes beyond simple detection by adding an embedded measurement layer that calculates the length and distance of targets, providing actionable data for inspections. This system integrates these features into a low-cost, modular ROV, filling a key gap between underwater detection algorithms and practical inspection needs. Also, it provides real-time perception and measurement for intervention frameworks, handling crack detection, length estimation, and standoff distance calculations.

\section{System Design and Implementation} \label{sdi}
The design philosophy of HyDRA Scorpion prioritizes maneuverability, modularity, and survivability in harsh underwater environments. This section details the system architecture, from its high-level operational principles to the specific implementation details of the prototype.

\subsection{System Overview}

\begin{figure}[ht] 
\centering 
\includegraphics[width=0.45\textwidth]{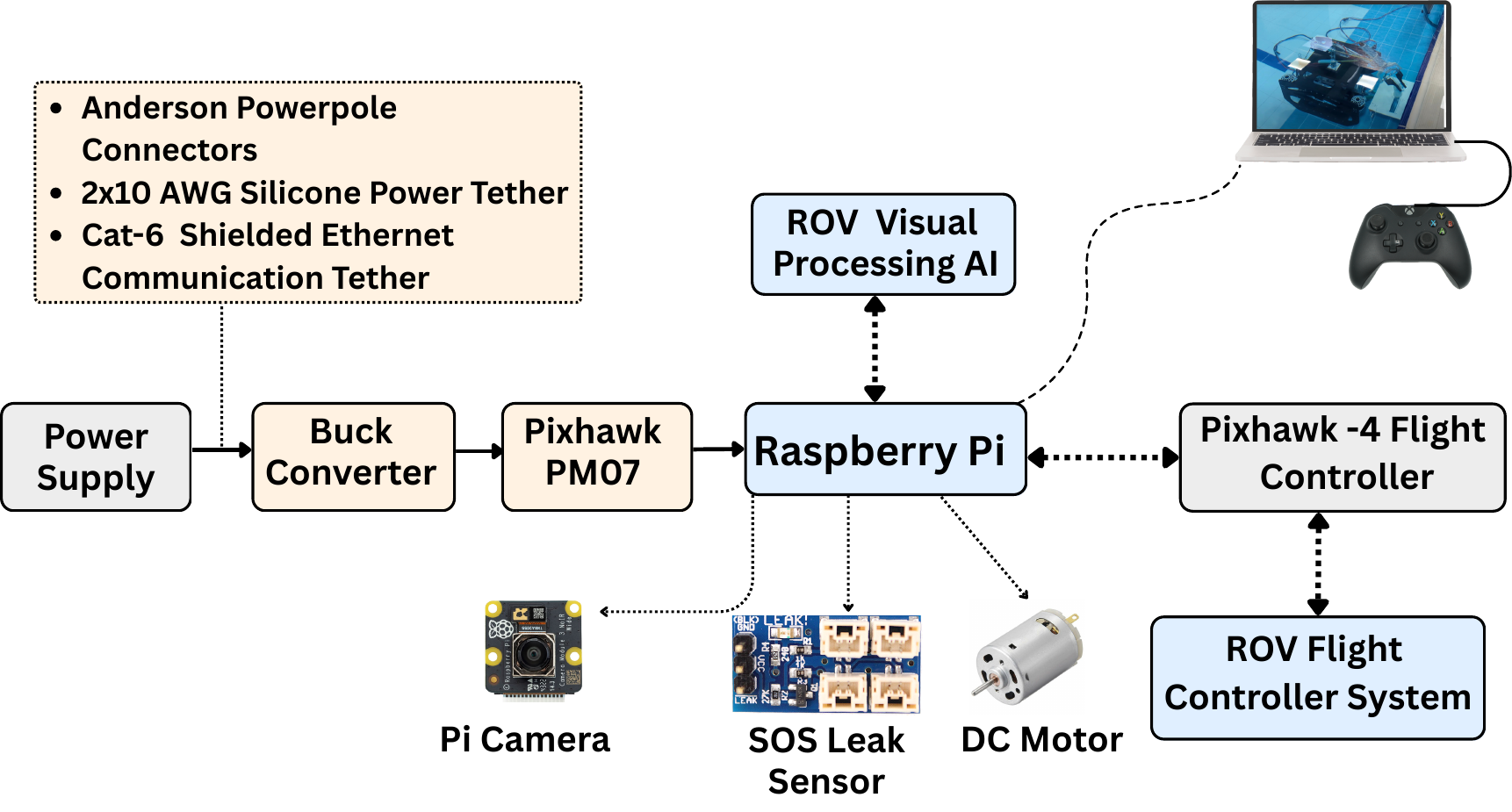} 
\vspace{-1em}
\caption{HyDRA Scorpion system overview.}
\label{fig:system_architecture}
\end{figure}

HyDRA Scorpion is a tethered, 6-Degree-of-Freedom (6-DoF) ROV designed for multi-purpose underwater intervention. The system architecture, depicted in Fig.~\ref{fig:system_architecture}, is divided into the physical vehicle and a topside operator station connected via a custom power-and-data tether. The system employs a hierarchical control structure. A Raspberry Pi serves as the high-level computer, managing visual AI processing and peripheral sensor data. It communicates with a Pixhawk 4, which acts as a real-time, low-level flight controller dedicated to vehicle stability and motor control. Power flows from a topside supply through onboard buck converters and a PM07 module to the various subsystems.

A central, 6-inch acrylic pressure vessel, fabricated in-house, houses the core electronics and is mounted within an open-frame chassis that supports all propulsion, manipulation, and sensory payloads. Power is delivered from the surface at 48V to minimize transmission losses and converted onboard. The ROV's vectored eight-thruster configuration provides full 6-DoF authority, enabling precise and agile maneuvering. A key feature of the architecture is the hierarchical computing model, which separates high-level perception and communication tasks from low-level, real-time vehicle control. This ensures robust and stable flight performance, even while running computationally intensive vision modules. The ROV's modularity is a core design principle, allowing for the integration of distinct operational capabilities that function like different "modes" of operation. These capabilities transform the ROV from a simple teleoperated camera into an intelligent tool for scientific measurement, environmental awareness, and complex intervention. Primary operational capabilities include object detection, in situ scientific measurement, 3D environmental mapping, and intervention.

\subsubsection{Object Detection} \label{Object_Detection}
In this capability mode, the ROV leverages its hierarchical computing architecture and advanced vision software to act as an intelligent observer. The Raspberry Pi 5 runs real-time object detection models on the live video feed. This enables the system to autonomously identify, classify, and track objects of interest, ranging from marine life, such as jellyfish and plastic bottles,  to mission-specific markers. Detections are overlaid on the operator's GUI, providing enhanced situational awareness and automating the data logging process. This mode is crucial for environmental monitoring, search and rescue operations, and ensuring operational safety by automatically identifying potential hazards and sensitive fauna.

\subsubsection{In-Situ Scientific Measurement} \label{In_Situ}
For tasks that require quantitative data, the ROV can operate in measurement mode. This capability combines a YOLO object detector with a calibrated pixel-to-centimeter ratio to perform accurate, non-contact measurements of underwater targets' length and distance. By calibrating the video feed with a reference object of known size, the operator can select points on an object of interest (e.g., a shipwreck artifact or coral formation) to obtain its real-world dimensions with high precision. This mode effectively transforms the ROV into a remote-controlled underwater caliper, enabling the collection of scientific data and structural inspection without the need for physical contact.

\subsubsection{3D Environmental Mapping}\label{3D}
This mode uses the precise 6-DoF control and the dual-manipulator system of the vehicle to interact with and map its environment. Using its 360° photosphere generation capability, the ROV can perform a controlled yaw maneuver at a point of interest to create a high-resolution, immersive 3D visualization of the surrounding area. This provides a comprehensive site overview for mission planning and post-dive analysis. Subsequently, the operator can utilize the 4-DoF "Stinger" arm for complex intervention tasks, such as component replacement, sample retrieval, or debris clearance, guided by a detailed environmental map and multiple camera views. This powerful combination of mapping and manipulation allows the ROV to perform tasks that would otherwise be difficult or impossible for human divers.

\subsection{Prototype Design}
The physical prototype of the HyDRA Scorpion was constructed with specific components and design choices forged through extensive testing and iteration. This subsection details the implementation of the mechanical and electrical systems.

\subsubsection{Mechanical Subsystems}
\begin{figure}[ht]
    \centering
    \includegraphics[width=0.5\textwidth]{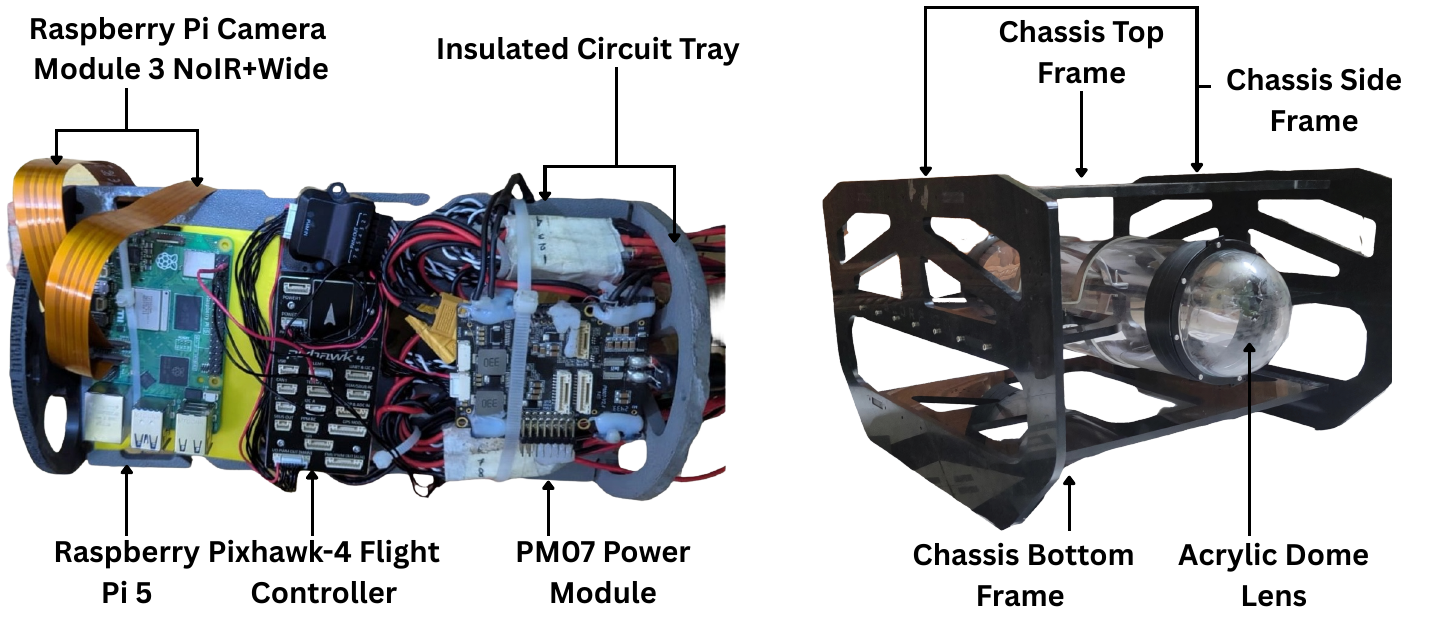}
    \vspace{-1.5em}
    \caption{Mechanical structure of HyDRA-Scorpion.}
    \label{fig:Mechanical_HyDRA}
\end{figure}

The mechanical assembly of the HyDRA Scorpion integrates a custom-made chassis, a central pressure housing, and multiple payloads into a cohesive system, as shown in Figure \ref{fig:Mechanical_HyDRA}. The design prioritizes modularity and field serviceability.

\paragraph{Frame and Hydrodynamics}
The chassis is fabricated from 10mm carbon black reinforced high-density polyethylene (HDPE) sheets. This material was selected over traditional aluminum alloys for its excellent corrosion resistance in saltwater, its near-neutral buoyancy (95\% of water density), which significantly simplifies vehicle trimming, and its environmental recyclability. The frame features a modular, open-structure design that provides high structural rigidity while minimizing hydrodynamic drag and allowing unrestricted water flow to the eight T200 thrusters.

\renewcommand{\arraystretch}{1.3}
\begin{table}[htbp]
\centering
\caption{Selected Mechanical elements for design}
\label{tab: Mechanical elements}
\begin{tabular}{lll}
\hline
 & \textbf{Materials} & \textbf{Specification} \\ \hline
Chassis & HDPE & 580 mm x 497 mm x 280 mm \\ 
\begin{tabular}[c]{@{}l@{}}Waterproof Circuit \\ Chamber\end{tabular} & \begin{tabular}[c]{@{}l@{}}Acrylic and \\ Aluminum Alloy\end{tabular} & 160 mm (Diameter) x 500 mm \\ 
Optical Dome Lens & Acrylic & 160 mm (Diameter) \\ 
Propeller & Mold Plastic & N/A \\ 
\begin{tabular}[c]{@{}l@{}}4-DoF Underwater \\ Manipulator\end{tabular} & \begin{tabular}[c]{@{}l@{}}Aluminum and \\ Brass\end{tabular} & \\ 
Tether Power & Silicone Wire & 2 x 10 AWG x 50 meter \\ 
Screw & Stainless Steel & N/A \\ 
\textbf{Total Weight} & 17.6 Kg & \\ \hline
\end{tabular}
\end{table}

\paragraph{Pressure Housing and Waterproofing}
Robust waterproofing presented a critical engineering challenge that we addressed through iterative design. All primary electronics are housed within a 6-inch diameter, 500mm acrylic cylinder. This design choice was the result of significant prototyping; an early version built with locally sourced aluminum sheets proved too heavy and suffered from recurring seal failures. Budget and sourcing constraints led us to reverse engineer a sample unit. The final internal enclosure was made from a locally available acrylic pipe and sealed with custom-machined end caps made of shipyard-grade aluminum. The crucial seal between the acrylic tube and the end caps is achieved using a double O-ring radial seal on each cap. The quality of these O-rings was found to be paramount; initial tests with local O-rings failed, necessitating the use of high-quality imported O-rings to provide a reliable seal. All cables penetrate the enclosure through these end caps via compression-gland connectors, which ensure a watertight pass-through while providing essential strain relief. This same waterproofing principle of sealed acrylic enclosures with O-rings is applied to protect subsystem components, including the manipulator motors and the two Raspberry Pi cameras. We validated the integrity of the final ROV through both simulated pressure tests and real-world deployments. In the lab, the housing was pressure tested to 4 bar, the equivalent of an approximate 300-meter water depth. This test confirmed a leak-free operation without moisture ingress. The ROV also operated successfully in a wide range of conditions. The deployments included an 18-foot-deep water tank and a 10-foot-deep natural waterway (canal).

\renewcommand{\arraystretch}{1.3}
\begin{table}[]
\centering
\caption{Selected Electrical components for design}
\label{tab:Electrical elements}
\begin{tabular}{ll}
\hline
\textbf{Component} & \textbf{Model \& Specification} \\ \hline
Flight Controller        & Pixhawk 4                          \\ 
Main Processing Unit     & Raspberry Pi 5 (8 GB)              \\ 
Main Camera              & Raspberry Pi Camera 3 NoIR         \\ 
Wide Angle Camera        & Raspberry Pi Camera 3 NoIR Wide    \\ 
Thruster                 & ROVMAKER T200 (7.1 kg/5.5 kg)      \\ 
ESC                      & ROVMAKER ESC 35A                   \\ 
Power Management Unit    & Pixhawk PM07                       \\ 
Manipulator Motor        & 12VDC 22 mm Planetary Gear Motor   \\ 
Tether Communication     & CAT-6 Ethernet (50 m)              \\ 
ROV Controller           & Gamesir Nova Light                 \\ 
Manipulator Controller   & Logitech F310                      \\ \hline
\end{tabular}
\end{table}

\begin{figure}[ht]
    \centering
    \includegraphics[width=0.6\linewidth]{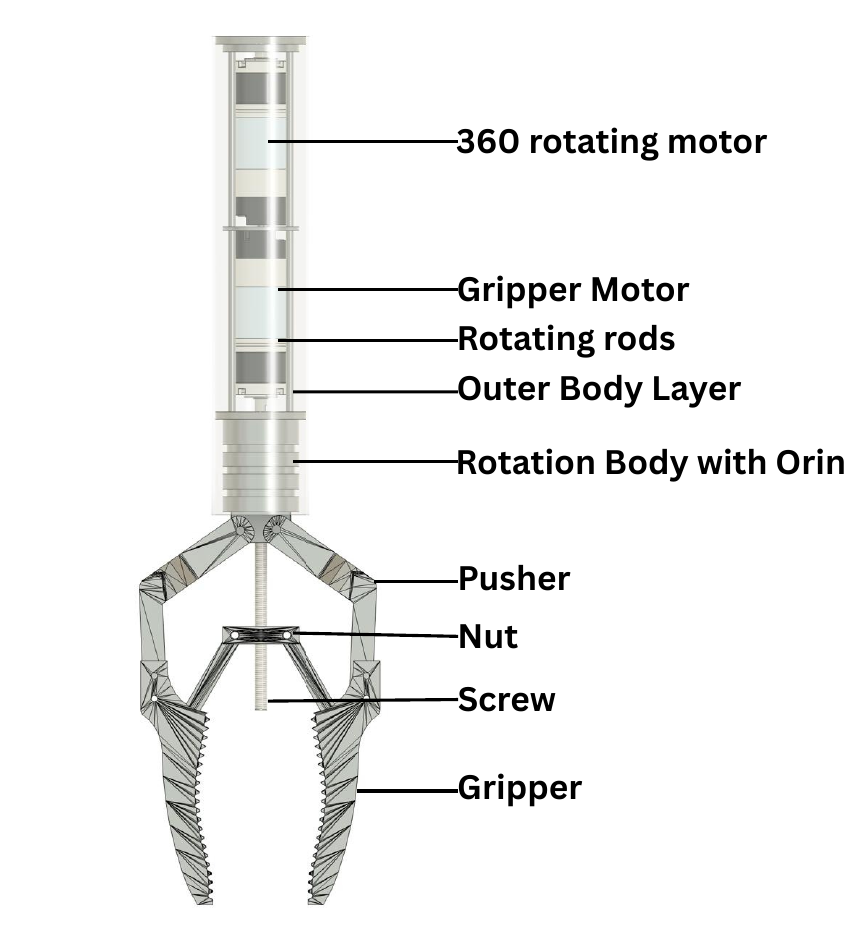}
    \vspace{-1em}
    \caption{Cross-section of the primary manipulator, detailing the coaxial dual-motor drive system and the lead-screw end-effector mechanism.}
    \label{fig:manipulator_overview}
\end{figure}

\paragraph{Manipulator Design}
The ROV's primary manipulator was designed and fabricated entirely in-house to meet the dual requirements of robust gripping and unlimited yaw rotation, a capability not found in affordable commercial options. The end effector's design is bioinspired by the dexterity of a shrimp's appendages. A significant engineering challenge was to allow the entire gripper assembly to rotate continuously 360 degrees without entangling the power and control wires for the gripper motor. We developed a novel coaxial drive mechanism to solve this problem, detailed in Figure~\ref{fig:manipulator_overview}. The arm's core is a cylindrical housing containing two distinct DC motors. The upper motor, the 360 Rotating Motor, is responsible for the yaw rotation of the entire end effector assembly. To solve the wire entanglement problem, a repurposed gear set from a cordless drill was integrated. This gear set transfers rotational torque to the Rotation Body while allowing the gripper motor's wires to remain stationary, providing a robust, low-cost solution. The lower Gripper Motor drives the gripping action. This motor shaft connects directly to a central lead screw. As the screw rotates, it drives a threaded Nut linearly along its vertical axis. This nut is mechanically linked to a set of Pushers, which translate the linear motion into the angular movement of the two opposing serrated Gripper jaws. This design efficiently converts motor rotation into a strong and precise gripping force. A simpler single-motor secondary gripper is also mounted on the vehicle to assist the main arm.

\paragraph{Dual Manipulator System} 
To perform complex intervention tasks, Scorpion is equipped with two manipulators. The primary arm, named "Stinger," is a 4-DoF manipulator driven by 12V Planetary Gear DC motors and BTS7960 drivers, utilizing lead-screw actuation for precise motion. The secondary gripper, called "Pincer," is a simpler single-motor mechanism. All motors are waterproofed in sealed acrylic compartments.

\subsubsection{Electrical and Computing Architecture}
\begin{figure}[ht]
    \centering
    \includegraphics[width=0.44\textwidth]{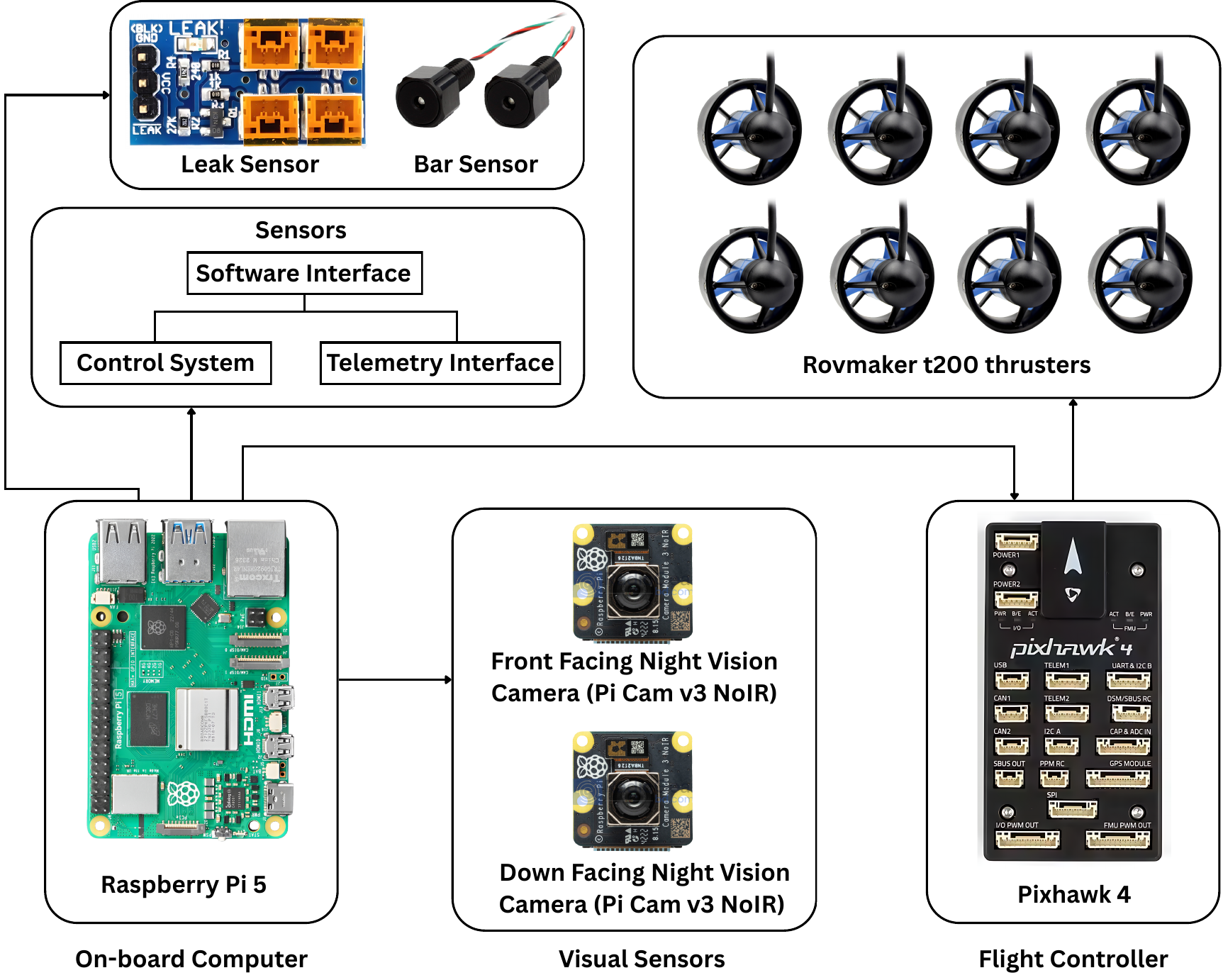}
    \vspace{-.5em}
    \caption{Electronics design of HyDRA-Scorpion}
    \label{fig:electronics}
\end{figure}

Electrical and computer systems are designed to provide reliable power distribution and separate high-level perception tasks from real-time vehicle control. The complete electronics design, showing the interconnection of all major components, is illustrated in Figure \ref{fig:electronics}.

\paragraph{Power Distribution and Tether} 
A topside 48V DC power supply delivers power down a custom tether, which merges two 10 AWG cables for power and a single CAT-7 S/FTP Ethernet cable. CAT-7 was specifically chosen not for bandwidth but for its superior shielding, which protects the data link from thruster-induced EMI. Power enters the ROV via an Anderson Powerpole SBS50 connector and is protected by a 30A LittleFuse JCASE fuse. Onboard, four parallel 48V-to-12V, 30A DC-DC buck converters provide high-current 12V power. For sensitive 5V electronics, multiple LM7805 voltage regulators with heatsinks are used in parallel to provide a stable power supply.

\paragraph{Computing and Sensory Suite} 
The ROV's hierarchical computing architecture consists of a Raspberry Pi 5 with 8 GB of RAM for high-level tasks and a Pixhawk 4 flight controller for real-time control. The sensor suite includes a BMP388 for internal pressure/temperature, a Blue Robotics Bar02 sensor for water depth, and an SOS Leak Sensor. For vision, the ROV is equipped with two Raspberry Pi Camera Module v3 NoIR units (one standard, one wide-angle), selected for their superior low-light sensitivity.

\subsection{Software and Control Systems}
The software stack is built on Python, utilizing PyQt6 for the GUI and OpenCV for perception. A client-server model over UDP/TCP handles communication.

\subsubsection{GUI and Data Management}
The operator station provides comprehensive control via a gaming joystick and real-time telemetry on an 18-inch monitor. The PyQt6 GUI displays video feeds, sensor data (depth, temperature, leak alerts), and thruster status. The pilot can switch between "Constant Thrust Mode" for steady transit and "Incremental Thrust Mode" for gradual acceleration. Communication is handled via socket programming, with hardware managed by RPi.GPIO and video are streamed using GStreamer. All mission data is logged to time-synchronized CSV files.

\subsubsection{Control and Thrust Allocation}
The vehicle's motion is governed by a cascaded PID control structure running on the Pixhawk. The desired wrench \(\boldsymbol{\tau}_d\) from the controllers and operator joystick is translated into motor commands \(\mathbf{f}\) by solving a constrained, weighted least-squares optimization problem, as shown in Eq.~\ref{eq:control}.
\begin{equation}
\min_{\mathbf{f}} \|\mathbf{W}(\mathbf{B}\mathbf{f}-\boldsymbol{\tau}_d)\|_2^2 \quad \text{s.t.} \quad f_i \in [f_i^{\min}, f_i^{\max}].
\label{eq:control}
\end{equation}
Here \(\mathbf{W}\) is a diagonal matrix for weighting control axes, and saturation is handled via anti-windup logic.

Algorithm 1 illustrates our proposed controller step by step.
\\
\rule{9cm}{0.03cm}
\textbf{Algorithm \label{algo2}} Incremental Station-Keeping Controller.\\
\hspace{0.05cm}
\rule{9cm}{0.03cm}
\\[.05cm]
\textbf{Input:}
\newline
Pose $(x,y,z,\phi,\theta,\psi)$, body rates, hold setpoint, user feed-forward.
\newline
\textbf{Output:}
\newline
Actuator commands $\mathbf{f}$ with anti-windup compensation.
\newline
\textbf{Method:}
\begin{enumerate}
  \item Read vehicle pose $(x,y,z,\phi,\theta,\psi)$ and body rates.
  \item Compute position and attitude errors $e$ relative to the hold setpoint.
  \item Apply PID control on $e$ to compute the desired wrench $\boldsymbol{\tau}_d$.
  \item Add user feed-forward terms for fine trim adjustments.
  \item Solve the control allocation problem to compute actuator commands $\mathbf{f}$.
  \item Apply $\mathbf{f}$ to ESCs with rate limiting to ensure smooth transitions.
  \item Perform anti-windup by back-calculating integral terms when saturation occurs.
  \item Return actuator commands $\mathbf{f}$.
\end{enumerate}

\subsection{Vision-Based Mission Capabilities}
A key innovation of the Scorpion platform is its suite of software modules that leverage computer vision to provide capabilities beyond simple teleoperation.

\subsubsection{AI-Based object Detection Module}
\begin{figure}[ht]
  \centering
   \includegraphics[width=0.30\textwidth]{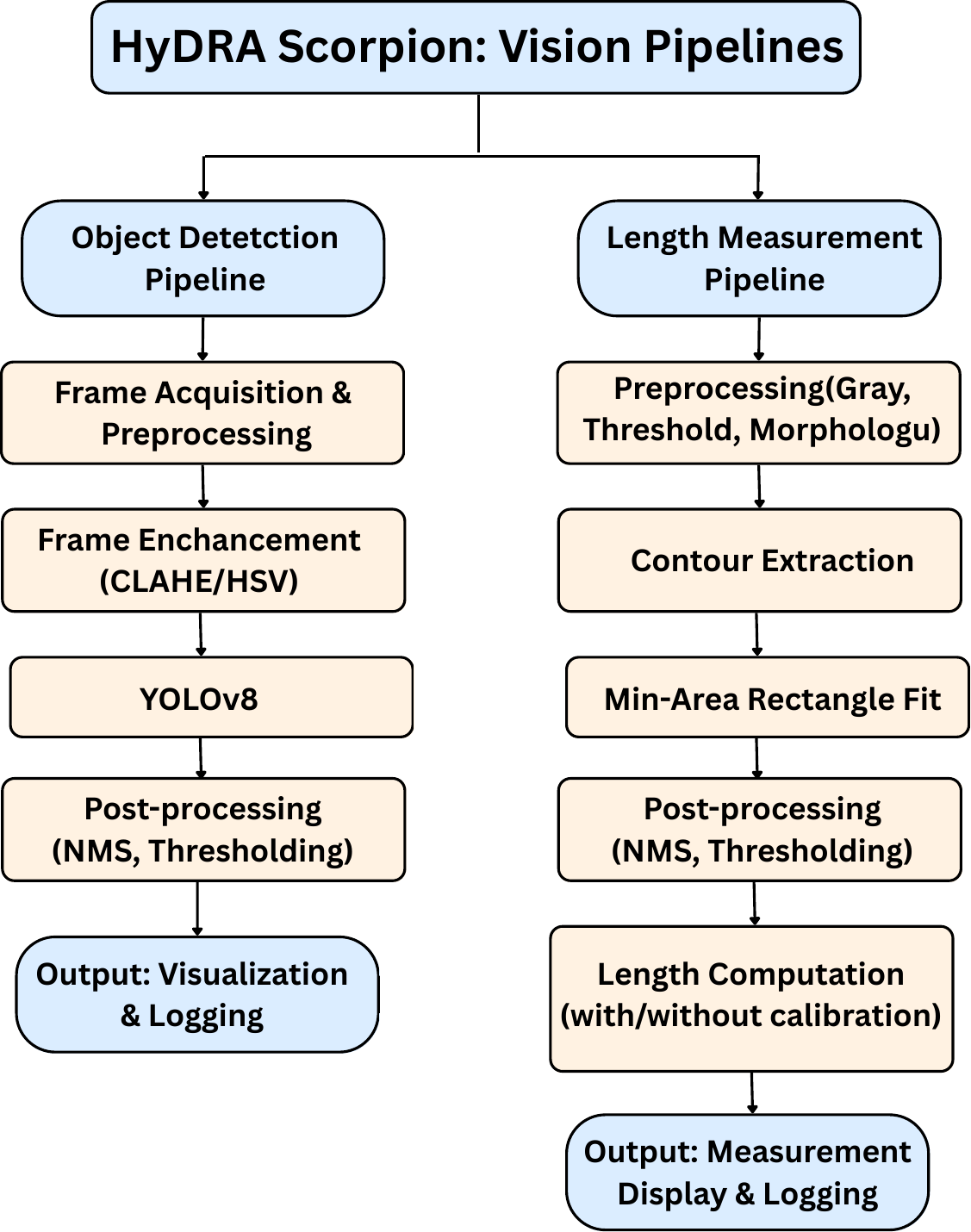}
   \vspace{-.5em}
  \caption{Vision Pipelines of object detection and length measurement}
  \label{fig:vision_pipeline}
\end{figure}

To enhance mission safety, support ecological research, and automate inspection tasks, a versatile AI-based detection module was integrated into the ROV. While initially focused on jellyfish, the system was expanded to recognize a wide range of underwater classes relevant to both environmental monitoring and industrial inspection. A YOLOv8 model \cite{reis2023real} was trained on a curated dataset of underwater images and deployed on the onboard computer using a TensorRT-optimized engine for efficient inference. As shown in the vision pipeline Figure~\ref{fig:vision_pipeline}, each camera frame is preprocessed before being fed to the model, which outputs bounding boxes and confidence scores for any detected objects. These detections are overlaid on the operator's video feed and logged with timestamps, allowing the operator to identify marine life, locate debris such as nets and plastic, and recognize structural features, including shipwrecks.

\begin {comment}
\begin{figure}[H]
    \centering
    \resizebox{0.3\linewidth}{!}{
    \begin{tikzpicture}[
        node distance=7mm,
        box/.style={draw, rounded corners, minimum width=3.5cm, minimum height=1cm, align=center}
    ]
        \node[box, fill=purple!8] (raw) {Raw Frame};
        \node[box, below=of raw] (pre) {WB + CLAHE};
        \node[box, below=of pre] (net) {CNN Detector (YOLO)};
        \node[box, below=of net] (post) {NMS + Filter};
        \node[box, below=of post] (viz2) {Overlay + Log};
        \draw[->] (raw) -- (pre);
        \draw[->] (pre) -- (net);
        \draw[->] (net) -- (post);
        \draw[->] (post) -- (viz2);
    \end{tikzpicture}
    }
    \caption{Vision processing pipeline.}
    \label{fig:vision_pipeline}
\end{figure}
\end{comment}

\subsubsection{Interactive Length Measurement}
This module allows the operator to perform accurate, real-world length measurements. The operator first calibrates the system by selecting two points on a reference object of known length (e.g., a PVC pipe). The software then calculates a scale factor (pixels/meter), which is used to convert user-selected pixel distances on a target object into a real-world measurement. Lens distortion correction and sub-pixel interpolation are employed to enhance accuracy.

\subsubsection{360° Photosphere Generation}
For comprehensive site visualization, this module stitches images from a slow yaw maneuver into an interactive 360° photosphere. An OpenCV pipeline uses the ORB algorithm for feature detection, a FLANN matcher, and RANSAC for robust homography estimation. The aligned images are warped using a spherical projection and blended to create an equirectangular panorama, which is then rendered in the GUI using OpenGL for immersive inspection.

\subsection{Color and Shape-Based Marker Detection}
For identifying specific mission markers, such as a colored "T" shape, a computationally efficient OpenCV pipeline is used. The process begins with color segmentation by converting video frames to the HSV color space, which is robust against underwater lighting variations. A binary mask is created using a specific HSV threshold to isolate the target color.

To reduce noise from sensor artifacts or backscatter, this mask is refined using a series of morphological operations. Contours are then extracted from the cleaned mask to identify distinct object boundaries. Each contour is validated by analyzing its shape and aspect ratio to confirm it matches the geometric properties of a "T" pattern. Detections satisfying both color and shape criteria are highlighted on the operator's GUI, providing reliable real-time feedback.

\section{Results and Analysis} \label{ra}
We conducted an extensive series of controlled tests and field exercises. These experiments were designed to validate the performance of the Hydra Scorpion's core operational capabilities. First, we describe the datasets used in this section.

\subsection{Description of Datasets}
We have used the COCO dataset\cite{tong2023rethinking}, which is a well-known benchmark in computer vision. It includes 80 object categories, and each of them has more than 300K images, all are labeled with confidence supporting annotated bounding boxes, linearly arranged segmentation masks and openly distributed keypoints. We used the standard split of 118k training, 5k validation and 20k test images. We also used the Jelly data dataset of Roboflow containing 6,882 jellyfish images due to COCO dataset not contain specific jelly fish or jelly fish like creature. The dataset is split into 5,777 training images (84\%), 773 validation images (11\%), and 332 test images (5\%).

\subsection{Object Detection and Classification}
\begin{figure}[ht]
    \centering
    \includegraphics[width=0.40\textwidth]{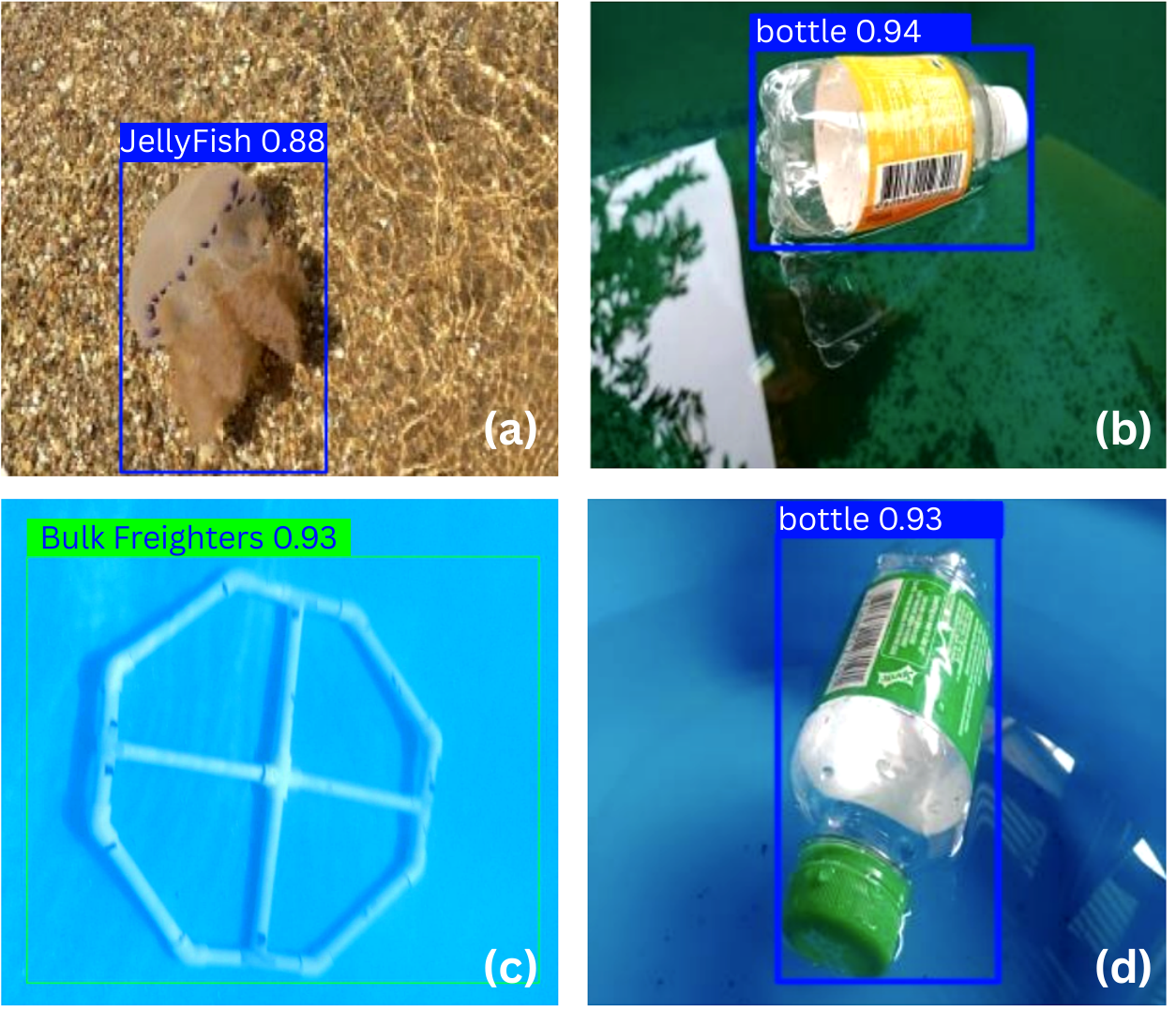}
    \vspace{-.5em}
    \caption{Sample output images from field tests of the AI vision system, showing real-time detection overlays on underwater targets. The model successfully identifies (a) marine life (jellyfish) against a complex, natural seabed, (b, d) man-made debris (bottles) in varying light and reflections, and (c) a mission-specific structural target (bulk freighter marker).}
    \label{fig:samples}
\end{figure}

Robust environmental perception is a primary challenge for any autonomous or semi-autonomous system. We evaluated the performance of the real-time object detection module to validate the ROV's capability as an intelligent observer. The YOLOv8 model was tested against a held-out, annotated dataset of underwater images. Key metrics were calculated to quantify the model's performance. These metrics included mean Average Precision (mAP@0.5) for general object detection and recall for specific industrial tasks. Qualitative tests were also conducted during field exercises. These tests observed the system's ability to identify targets in real-time. The object detection system performed with high accuracy. The YOLOv8 model achieved a mean Average Precision of 0.89 on the jellyfish dataset. It also achieved a recall of 88\% on the specialized crack detection task. During field exercises, the system successfully identified mission-critical markers and environmental features in real-time. Detections were correctly overlaid on the operator's GUI, as shown in Figure~\ref{fig:samples}. These results confirm the system's effectiveness for both environmental monitoring and industrial inspection.

\subsection{In-Situ Scientific Measurement}
The ROV's function as a remote measurement tool required validation. This process involved confirming the accuracy of its non-contact dimensioning algorithms against ground-truth data. The interactive length measurement tool was tested in a controlled pool environment. A reference object of known size first established a pixel-to-metric scale factor. The operator then used the tool to measure the dimensions of several other submerged objects. The lengths of these objects were precisely known. The percentage error between the ROV's measurement and the ground truth was calculated for each object. The system's color detection accuracy was also benchmarked against red, blue, and yellow markers.

\begin{figure}[ht]
    \centering
    \includegraphics[width=0.40\textwidth]{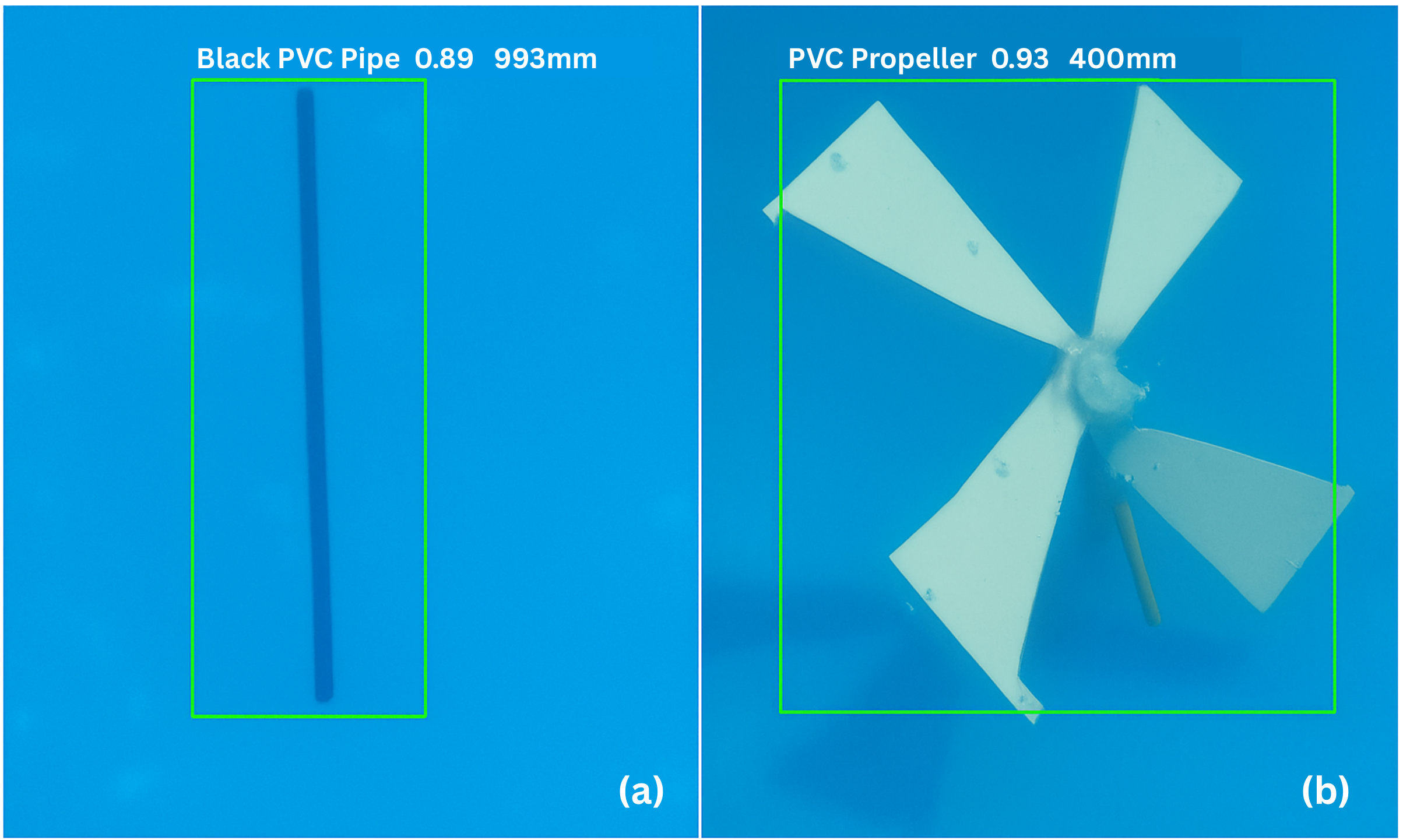}
    \vspace{-.5em}
    \caption{Sample outputs from the vision-based measurement system. The system detects and measures the dimensions of underwater objects in real-time, such as (a) a PVC pipe and (b) a propeller.}
    \label{fig:length_measurement_samples}
\end{figure}

The measurement capability proved to be highly accurate. The interactive length measurement tool demonstrated a consistent measurement error of less than 5\% across all test objects. Sample outputs from the measurement tool during these validation tests are shown in Figure~\ref{fig:length_measurement_samples}. The underlying color detection algorithm achieved an average accuracy of 94\%. A summary of these quantitative results is presented in Table \ref{tab:metrics}.

\begin{table}[ht]
    \centering
    \caption{Selected performance metrics from vision system validation.}
    \label{tab:metrics}
    \footnotesize
    \begin{tabular}{@{}lll@{}} 
        \toprule
        \textbf{Metric} & \textbf{Result} & \textbf{Conditions / Notes} \\
        \midrule
        Color Detection & 94\% Accuracy & Tested on Red/Blue/Yellow markers \\
        Length Measurement & < 5\% Error & Using a calibrated reference marker \\
        Jellyfish Detection & 0.89 mAP@0.5 & On a held-out test image split \\
        Crack Detection & 88\% Recall & On pre-annotated painted panels \\
        \bottomrule
    \end{tabular}
\end{table}

\subsection{3D Environmental Mapping and Intervention}
The final validation focused on the integrated performance of the vehicle's mapping and manipulation capabilities. A primary challenge is performing a stable maneuver to capture high-quality images for stitching. This test also needed to demonstrate the utility of the resulting map for a subsequent intervention task.

\begin{figure}[ht]
    \centering
    \includegraphics[width=0.40\textwidth]{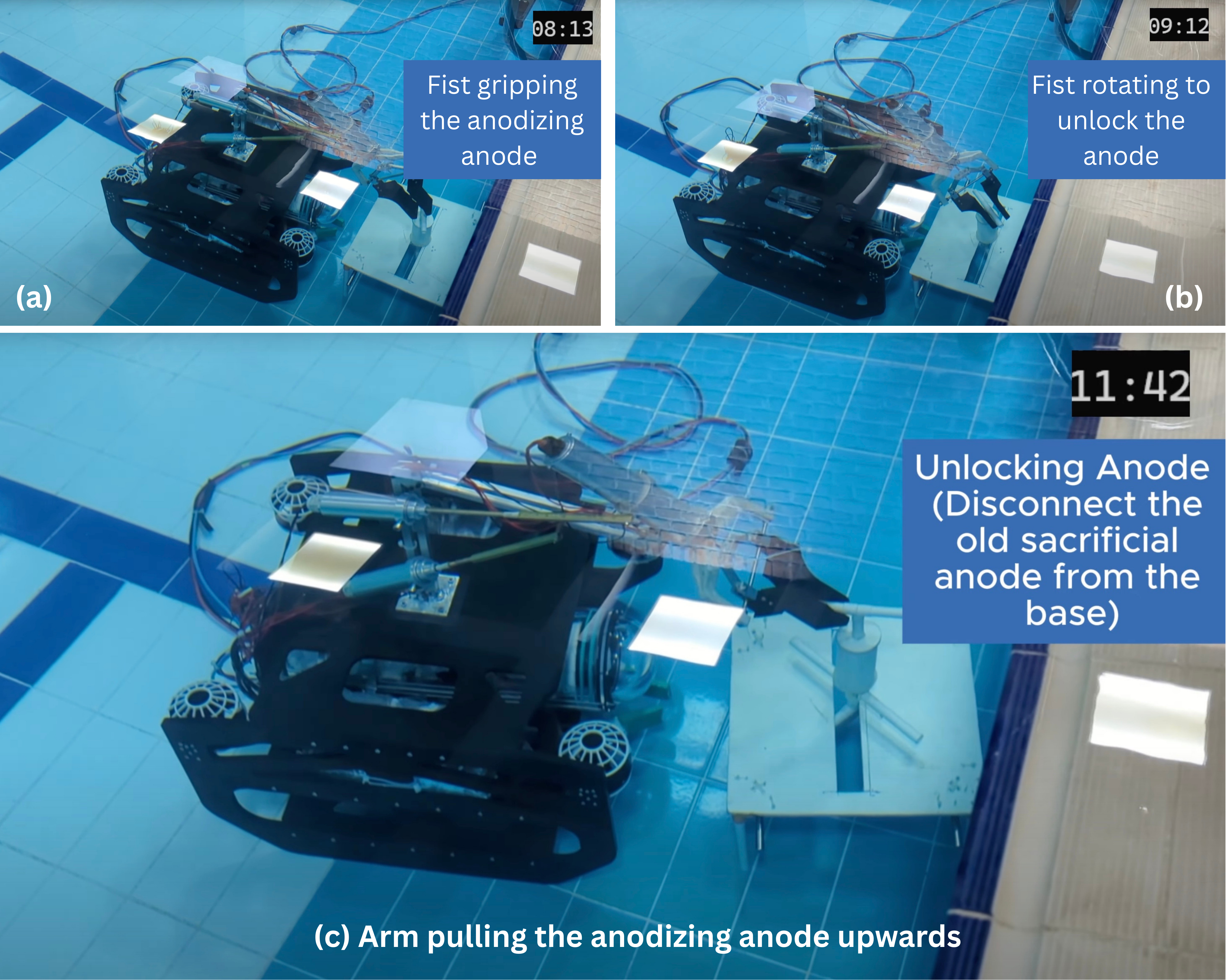}
    \vspace{-.5em}
    \caption{A full operation of the Manipulator Arm in the test bed}
    \vspace{-1em}
    \label{fig:manipulator_operation}
\end{figure}

The ROV was commanded to perform a slow, 360-degree yaw maneuver. It maintained a fixed position in the center of a test area populated with various objects. The ground station processed the captured images to generate a panoramic photosphere. Subsequently, the operator used this photosphere to plan a path to one of the objects. The operator then performed a mock anode replacement task with the primary manipulator. The vehicle's precise station-keeping control enabled the capture of a stable image sequence, resulting in a clear, seamless 360° panoramic photosphere with no significant stitching artifacts. Using this map, the operator successfully planned and executed the intervention task. The mock anode replacement, a multi-step process involving gripping, rotating to unlock, and retracting the anode, was completed well within the allotted time. The key stages of this successful manipulator operation are documented in Figure~\ref{fig:manipulator_operation}. This end-to-end test validated the powerful synergy between the vehicle's precise control, environmental mapping, and manipulation capabilities.

\section{Discussion}
\subsection{Design and Implementation Efficiency}
The total cost of the HyDRA Scorpion ROV system is approximately \$4,542.98 as detailed in Table~\ref {tab:cost_table}. Its complement of eight T200 thrusters (\$2,058.61), Raspberry Pi 5 (\$126.63), Pixhawk 4 flight controller (\$120.01), Raspberry Pi camera (\$74.91), and assortment of specialized sensors (barometer, BMP390, SOS leak detector, and current sensor) allow for both high-level control as well as faithful underwater missions. Additional production costs such as the creation of the frame (\$369.78), creation of the arm (\$378.16), and material for the chassis (\$263.06) are significant factors in the durability and structural integrity of the platform. Accessories and tools, such as a 3D printer, soldering tools, power supply, buck converters, and wire, were also factored into the overall investment to provide for prototyping and subsequent maintenance. This cost-effectiveness, when weighed against the system's operational effectiveness, presents the ROV as a practical solution for underwater exploration and task execution. The system's design enables scalability and rapid deployment, while remaining flexible for various mission scenarios at an economical price point compared to most commercial underwater systems.

\begin{table}[ht]
    \centering
    \caption{Total design and implementation cost of HyDRA Scorpion}
    \label{tab:cost_table}
    \footnotesize
    \begin{tabular}{@{}lc@{}}
        \toprule
        \textbf{Component Name} & \textbf{Price in USD} \\
        \midrule
        T200 Thrusters                 & \$2,058.61 \\
        Raspberry Pi 5                 & \$126.63 \\
        Arducam Noir wide angle camera & \$74.91 \\
        Pixhawk 4                      & \$120.01 \\
        bt60 motor driver              & \$25.66 \\
        Joystick                       & \$131.42 \\
        Bar sensor                     & \$84.86 \\
        bmp390                         & \$10.92 \\
        sos leak sensor                & \$35.03 \\
        Current sensor                 & \$1.23 \\
        60 feet 10 awg silicon cable   & \$123.30 \\
        Anderson connector             & \$13.00 \\
        Anderson connector pin         & \$4.00 \\
        Power supply (48V 30A)         & \$328.63 \\
        Buck converter 48V to 12V, 30A & \$53.21 \\
        Multimeter                     & \$9.40 \\
        Soldering iron with stand      & \$10.27 \\
        Vero board                     & \$0.28 \\
        Body Manufacturing             & \$369.78 \\
        Arm Manufacturing              & \$378.16 \\
        3D printer                     & \$320.61 \\
        Chassis Material               & \$263.06 \\
        \midrule
        \textbf{TOTAL} & \textbf{\$4,542.98} \\
        \bottomrule
    \end{tabular}
\end{table}


\section{Conclusion} \label{con}
We presented the design, implementation, and validation of HyDRA Scorpion, a 6-DoF ROV designed for complex underwater inspection and intervention tasks using a low-cost, in-house fabrication methodology. The final design is the culmination of a rigorous iterative development process that successfully navigated significant sourcing and engineering challenges. The vehicle's architecture combines a vectored eight-thruster configuration for precise 6-DoF control with a novel, bio-inspired manipulator featuring a coaxial drive for unlimited rotation. The performance of the system was validated through extensive experiments, which confirmed the integrity of the pressure housing at 4 bar, the control system's ability to maintain stationkeeping with an error of less than ±0.15 m, and the high precision of its vision-based modules, including a YOLOv8 object detector that achieved 0.89 mAP. HyDRA Scorpion demonstrates a successful framework for developing capable, vision-enhanced robotic systems with limited resources, providing a robust platform for future research in autonomous underwater operations.

Future work will focus on expanding the vehicle's autonomy and sensory capabilities. Key developments will include the implementation of SLAM and onboard path planning algorithms. We will also broaden the AI detector's scope to multi-class species and debris recognition through further dataset curation. Mechanically, we plan to upgrade the pressure housing to achieve a depth rating of \SI{100}{m}. Finally, the project will investigate the integration of stereo cameras or structured light sensors. These additions will improve metric estimation and support a new data-sharing pipeline for georeferenced environmental monitoring.

\bibliographystyle{IEEEtran}
\bibliography{templete}


\end{document}